\documentclass[10pt,twocolumn,letterpaper]{article}

\usepackage[pagenumbers]{iccv} 

\definecolor{iccvblue}{rgb}{0.21,0.49,0.74}
\usepackage[pagebackref,breaklinks,colorlinks,allcolors=iccvblue]{hyperref}
\usepackage{multirow}
\usepackage[accsupp]{axessibility}
\def\paperID{15864} 
\def\confName{ICCV}
\def\confYear{2025}

\newcommand{\OURS}{SC-Lane }

\title{SC-Lane: Slope-aware and Consistent Road Height Estimation Framework for 3D Lane Detection}

\author{
Chaesong Park$^{1}$ \quad Eunbin Seo$^{2}$\thanks{This work was done when Eunbin was at Seoul National University.} \quad Jihyeon Hwang$^{1}$ \quad Jongwoo Lim$^{1}$\thanks{Corresponding author.} \\
$^{1}$Seoul National University \quad
$^{2}$Hyundai Motor Group \\
{\tt\small chase121@snu.ac.kr, eunbinseo@hyundai.com, zhyeon@snu.ac.kr, jongwoo.lim@snu.ac.kr}
}

\begin{document}
\maketitle
\begin{abstract}
In this paper, we introduce SC-Lane, a novel slope-aware and temporally consistent heightmap estimation framework for 3D lane detection. Unlike previous approaches that rely on fixed slope anchors, SC-Lane adaptively determines the fusion of slope-specific height features, improving robustness to diverse road geometries. To achieve this, we propose a Slope-Aware Adaptive Feature module that dynamically predicts the appropriate weights from image cues for integrating multi-slope representations into a unified heightmap. Additionally, a Height Consistency Module enforces temporal coherence, ensuring stable and accurate height estimation across consecutive frames, which is crucial for real-world driving scenarios.

To evaluate the effectiveness of SC-Lane, we employ three standardized metrics—Mean Absolute Error (MAE), Root Mean Squared Error (RMSE), and threshold-based accuracy—which, although common in surface and depth estimation, have been underutilized for road height assessment. Using the LiDAR-derived heightmap dataset introduced in prior work \cite{park2024heightlane}, we benchmark our method under these metrics, thereby establishing a rigorous standard for future comparisons. Extensive experiments on the OpenLane benchmark demonstrate that SC-Lane significantly improves both height estimation and 3D lane detection, achieving state-of-the-art performance with an F-score of 64.3\%, outperforming existing methods by a notable margin. For detailed results and a demonstration video, please refer to our project page: https://parkchaesong.github.io/sclane/

\end{abstract}    
\section{Introduction}
\label{sec:intro}

Accurate 3D lane detection from a camera is crucial for autonomous driving, but estimating road height remains a challenging problem. Existing 3D lane detection methods primarily follow two paradigms: Bird’s Eye View (BEV) feature-based methods and transformer-based methods, both of which struggle with complex road geometries.

BEV-based approaches \cite{wang2023bev, pittner2024lanecpp, zheng2024pvalane, chen2022persformer, Garnett20193dlanenet} extract features from front-view images and transform them into a BEV representation using Inverse Perspective Mapping (IPM). However, since IPM is highly sensitive to camera extrinsic variations, BEVLaneDet \cite{wang2023bev} introduces a virtual image plane to improve robustness against such changes. Alternatively, LaneCpp \cite{pittner2024lanecpp} incorporates a depth branch to enhance BEV projection. PVALane \cite{zheng2024pvalane} refines this approach by introducing a secondary sampling step of BEV features for improved lane estimation.

\begin{figure}[t]
    \centering
    \includegraphics[width=1\linewidth]{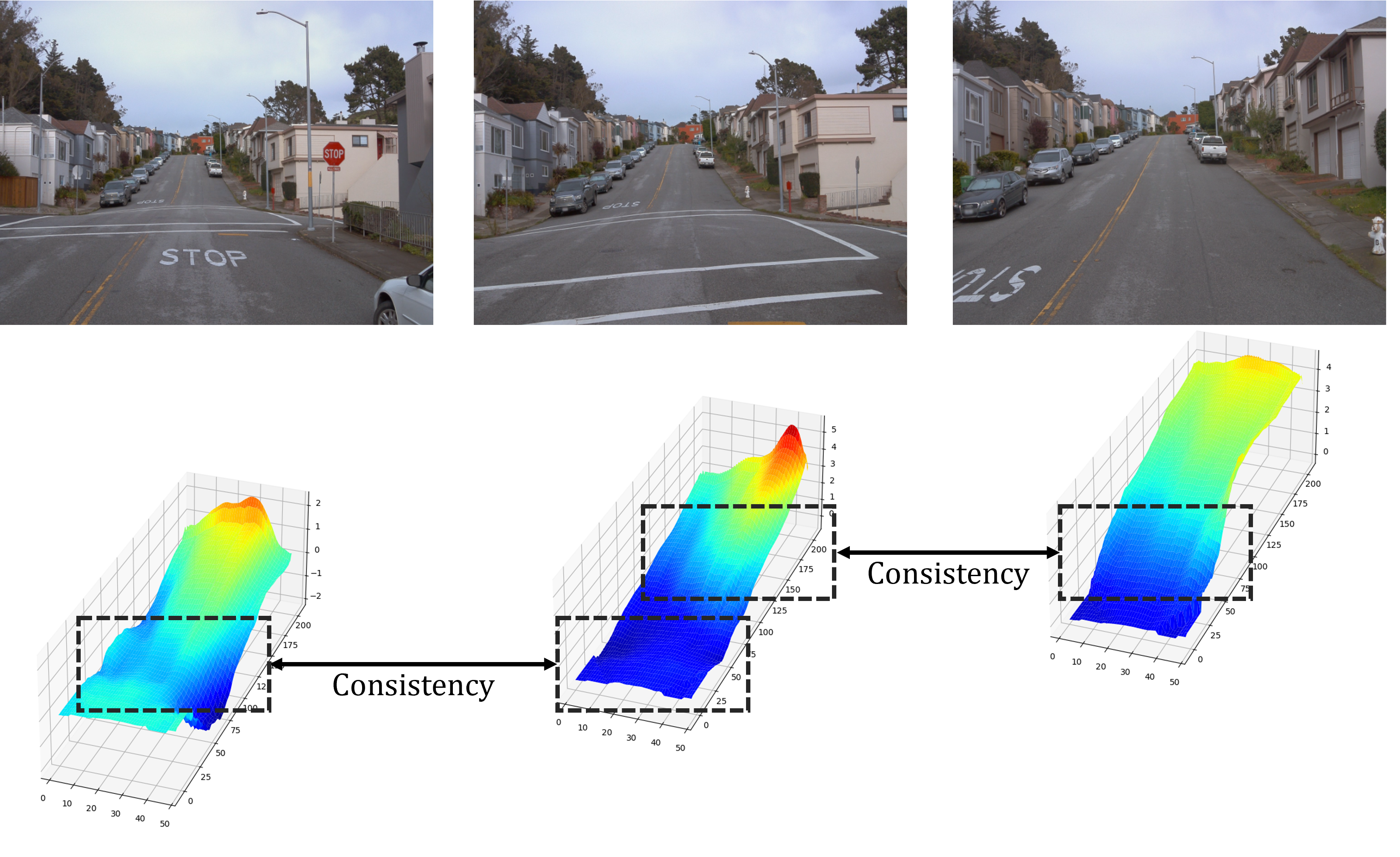}
   \caption{
Visualization of SC-Lane’s estimated height across consecutive frames.
The dashed black box highlights regions where height consistency should be preserved over time. Since the road surface is inherently static, the estimated heightmap should remain unchanged across time, aside from motion compensation due to vehicle movement.
The corresponding video is included in the supplementary materials.
   }
   \label{fig:heightmap}
\end{figure}

In contrast, transformer-based methods \cite{luo2023latr, liu2023petrv2, li2024repvf} directly operate on front-view features, avoiding explicit BEV transformation. These methods leverage 3D positional encodings to infer spatial information in predicting 3D lanes. PETRv2 \cite{liu2023petrv2} encodes potential 3D coordinates of the pixels into its positional encoding, while LATR \cite{luo2023latr} imposes a 2-DOF constraint to estimate the ground plane and reproject it into the image as an additional positional encoding. 

While these approaches have improved lane detection, they fail to address the challenge of accurate and temporally stable height estimation. HeightLane \cite{park2024heightlane}, a BEV-based method, introduced Multi-Slope Anchors to incorporate height information by sampling predefined slopes in the front-view feature space. The predicted heightmap is used implicitly as a positional encoding when constructing BEV features, to determine reference 2D pixels in a deformable attention module. However, this approach lacks adaptability, as it treats all slope features equally, without considering their relative importance in different road conditions. As a result, it struggles to generalize across diverse terrains. 

To overcome these limitations, we propose SC-Lane, a novel slope-aware and temporally consistent heightmap estimation framework. Unlike previous methods, SC-Lane dynamically determines the appropriate fusion of slope-specific height features, improving adaptability to complex road geometries. Our Slope-Aware Adaptive Feature module learns to blend multi-slope representations based on the input scene, enhancing height estimation accuracy and interpretability.

Furthermore, we introduce a Height Consistency Module that enforces temporal coherence, ensuring stable and consistent height predictions across consecutive frames (\cref{fig:heightmap}). 
Since road height should remain static relative to the ground, enforcing temporal consistency is particularly beneficial for height estimation. This concept has been explored in depth estimation \cite{godard2019digging, 8100182, ding2024towards}, where regularizing depth predictions over time improves robustness against dynamic objects. However, height estimation exhibits greater static properties than depth estimation, making it an even stronger candidate for temporal regularization. By leveraging this property, SC-Lane achieves more reliable height estimation, ultimately improving the accuracy of 3D lane detection.

Despite the increasing focus on 3D lane detection, standardized benchmarks for road height estimation remain underdeveloped. Most existing studies either rely on aerial imagery \cite{li2020height, mou2018im2height,zhang2025ts}, which is primarily designed for large-scale elevation mapping rather than road-level height estimation, or focus on robotic navigation \cite{chung2024pixel, yang2022real}, which operates in unstructured terrains rather than structured road surfaces. 
This lack of well-defined evaluation metrics has made it difficult to directly compare height estimation methods, limiting progress in this area. To address this gap, we introduce Root Mean Squared Error (RMSE), Mean Absolute Error (MAE), and threshold-based accuracy as standardized evaluation metrics for road height estimation. Using these metrics, we evaluate SC-Lane and baseline models, establishing a rigorous benchmark for future research and providing a more reliable framework for assessing height estimation methods.

\begin{figure*}[t]
    \centering
    \includegraphics[width=1\linewidth]{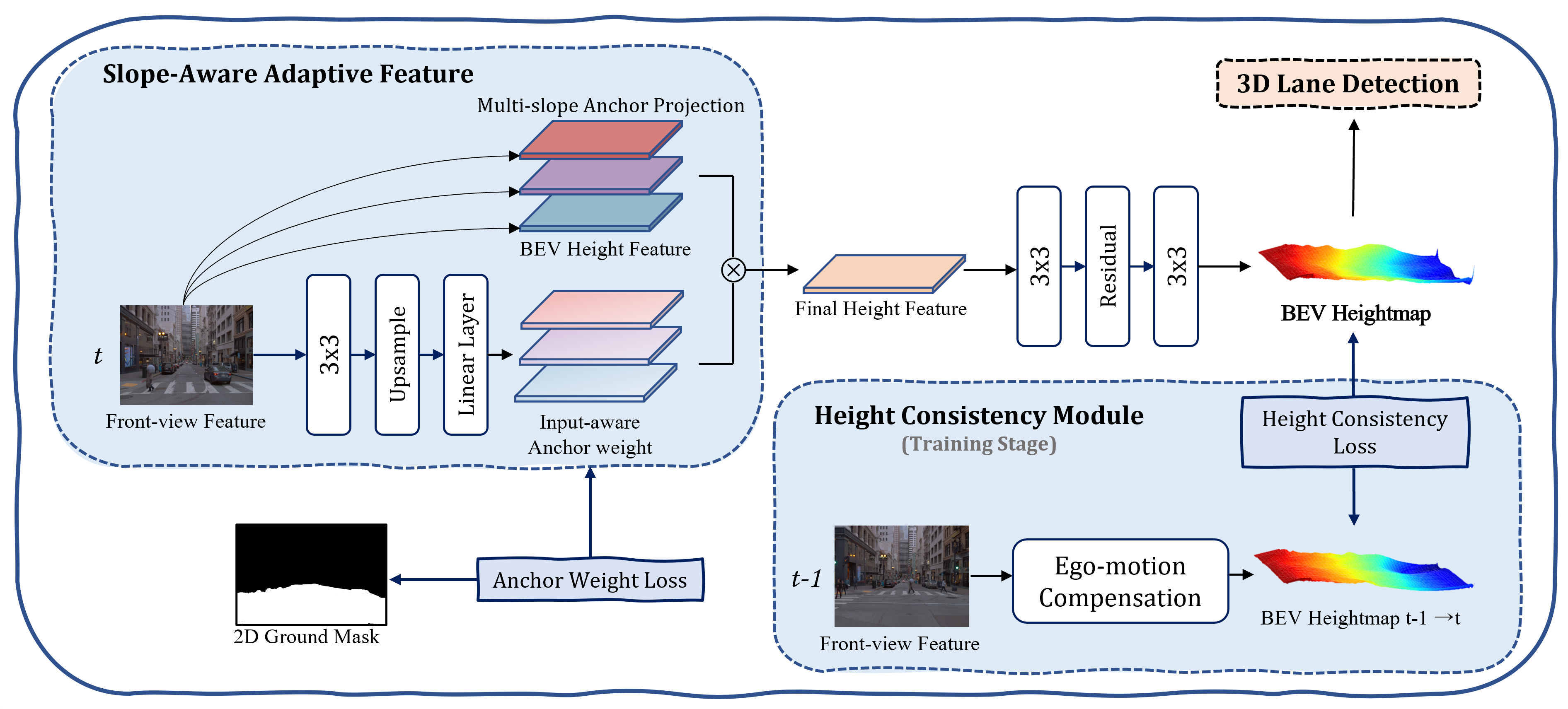}

   \caption{
\textbf{The overall architecture of SC-Lane}.
SC-Lane comprises two key components: the Slope-Aware Adaptive Feature module and the Height Consistency Module.
The Slope-Aware Adaptive Feature module adaptively fuses multi-slope representations by assigning weights to each slope-specific height features, enhancing height estimation accuracy across diverse road geometries.
The Height Consistency Module, applied during training, compensates for ego-motion by enforcing temporal coherence in height predictions across consecutive frames.
The final BEV heightmap, refined through these modules, serves as a geometric prior in 3D lane detection, ensuring improved lane localization and robustness in real-world autonomous driving scenarios.
   }
   \label{fig:architecture}
\end{figure*}

Our key contributions can be summarized as follows:
\begin{itemize}
    \item We introduce Slope-Aware Adaptive Feature to improve height estimation accuracy. Unlike previous methods that treat all slopes equally, our approach dynamically assigns adaptive weights to slope-specific features based on the input image. This allows the model to effectively utilize critical slope information, leading to significant performance gains, especially in undulating or steep road environments.
    \item We enforce temporal consistency for stable height estimation. Since road height is inherently static, enforcing temporal consistency across consecutive frames is essential. We introduce a temporal consistency loss that ensures smooth height predictions over time, improving both prediction reliability and overall lane detection accuracy.
    \item We establish a standardized benchmark for road height estimation by introducing three evaluation metrics: RMSE, MAE, and threshold-based accuracy. We also evaluate our method using these metrics, laying the foundation for future benchmarking in this field.
    \item SC-Lane achieves state-of-the-art performance in 3D lane detection. Our approach records an F-score of 64.3\% on the OpenLane benchmark, outperforming existing methods and demonstrating superior robustness in challenging driving conditions.
\end{itemize}
\section{Related Works}
\label{sec:RelatedWorks}

\subsection{Road Height Estimation}

Height estimation has been explored in various contexts, but explicit ground surface estimation remains relatively underdeveloped. Many studies focus on object-centric height estimation rather than modeling the road itself. For example, HeightFormer \cite{wu2024heightformer} integrates height estimation into occupancy prediction and 3D object detection, incorporating dynamic objects rather than directly estimating the ground. Additionally, its reliance on single-frame LiDAR data results in sparse ground-truth height annotations. Other studies \cite{zhang2025tssatmvsnet, li2020height, hou2023monocular} have addressed height estimation in aerial imagery, where heightmaps are inferred from a top-down perspective, resembling depth estimation rather than direct ground modeling.

In contrast, HeightMapNet \cite{qiu2024heightmapnet} and HeightLane \cite{park2024heightlane} specifically tackle ground height estimation. HeightMapNet uses a multi-camera setup for height estimation within an HD map reconstruction task. However, due to the absence of ground-truth height values, it discretizes height into fixed bins rather than regressing continuous values. HeightLane, on the other hand, predicts heightmaps from monocular images using multi-slope anchors, improving 3D lane detection. However, since height estimation is not its primary objective, it lacks comprehensive quantitative evaluation.

\subsection{Temporal Consistency}

Several prior depth estimation methods address consistency, ensuring that depth maps from adjacent frames align over time. One representative method involves exchanging information between consecutive frames to enforce smooth depth transitions. RollingDepth \cite{ke2024rollingdepth} estimates depth using short video snippets and applies an optimization-based global alignment process to maintain consistency. DepthCrafter \cite{hu2024depthcrafter} divides long video sequences into overlapping segments and aligns the scale and shift of depth distributions between segments to ensure smooth connectivity across frames. Furthermore, ST-CLSTM \cite{zhang2019exploiting} utilizes a CLSTM-based spatio-temporal modeling approach and applies an additional temporal consistency loss to enhance the smoothness of depth transitions between frames.

Another widely used approach is reprojection-based consistency. Monodepth2 \cite{godard2019digging} enforces temporal alignment by minimizing the photometric reprojection error during training, ensuring consistent depth predictions across frames. However, such approaches often include dynamic objects, making it difficult to guarantee that reprojection losses always guide the model towards better performance. To mitigate this issue, coarse-to-fine self-supervised DE training strategy \cite{moon2024from} excludes regions containing dynamic objects from the reprojection loss calculation, leading to more reliable depth learning.

Building on these advancements in depth estimation, our work specifically focuses on the road surface, which is largely static and thus more amenable to consistent geometric constraints. We introduce a Height Consistency Module that enforces alignment between height estimations across consecutive frames, leveraging the inherently stable nature of roads. This strategy results in more temporally coherent height maps, ultimately improving the performance of 3D lane detection in real-world driving scenarios.
\section{Methods}
\subsection{Overall Architecture}

The overall architecture of \OURS is illustrated in \cref{fig:architecture}. The input image is first processed through the feature backbone to extract front-view features. To capture diverse slope variations, multi-slope anchor projection is applied, following \cite{park2024heightlane}. However, rather than relying on predefined slopes, we propose a Slope-Aware Adaptive Feature module, which dynamically fuses slope-specific features based on the input image. This enhances the adaptability of height estimation and ensures robustness across varying road conditions.

To maintain sequential consistency in height predictions, we introduce the Height Consistency Module during training. This module enforces temporal stability by aligning consecutive heightmaps using ego-motion compensation.

The predicted heightmap is then utilized in the Height-Guided Spatial Transform, as proposed in \cite{park2024heightlane}, to refine BEV features. These BEV features are subsequently processed by a 3D lane head \cite{wang2023bev} to generate the final 3D lane predictions.

\subsection{Slope-Aware Adaptive Feature}

Since lanes are defined on the ground, extracting features related to the ground for height estimation was first introduced in HeightLane \cite{park2024heightlane}.
HeightLane proposed the use of multi-slope anchors, where each anchor samples projected 2D image features ${F}_{sampled}$ and concatenates them to generate BEV ground features.
However, this structure makes it challenging for the model to learn which slope features it should focus on the most.
To address this issue, we propose the Slope-Aware Adaptive Feature, which leverages image features to allow the model to predict the optimal focus among different slope-specific features.

Each slope anchor $a$ $(a = 1, \ldots, A)$ is defined as a 3D coordinate (X, Y, $\text{Z}_{a}$) at a specific (X,Y) location in the BEV grid, maintaining a fixed slope.
Projecting it into the 2D image space and sampling the image feature at the corresponding location to generate the BEV feature is as follows:
\[
    F^{a}_{sampled} = \mathcal{S}(\mathcal{P}(X, Y, Z_a), F_{img}),
\]
where $\mathcal{P}(X, Y, Z_a)$ is a function that projects the 3D coordinate $(X, Y, Z_a)$ into the camera view, and $\mathcal{S}(\cdot, F_{img})$ is an operation that samples the image feature $F_{img}$ at the projected 2D coordinate.

As illustrated in \cref{fig:architecture}, each slope anchor \(a\) produces a sampled BEV feature \(F^{a}_{\text{sampled}}\)$(\cdot)$.
To decide which anchor matters most at every BEV pixel \((X,Y)\), we predict a anchor weight \(\alpha_a(X,Y)\) from the backbone image feature:
a lightweight CNN upsamples the feature to BEV resolution, then a linear layer reshapes it to an \(A\)-channel tensor that is \emph{spatially} aligned with the anchor grid.
Applying a softmax along the anchor dimension yields a per-pixel probability distribution
\(\sum_{a=1}^{A}\alpha_a=1\).

The final height-aware feature is the probability-weighted blend of all anchor features,
\[
   F_{\text{SA}}(X,Y)=\sum_{a=1}^{A}\alpha_a(X,Y)\,F^{a}_{\text{sampled}}(X,Y).
\]

The anchor weights already act as pixel-wise confidences, so we convert them into a coarse height map
\[
   \mathcal{H}_{\text{conf}}(X,Y)=\sum_{a=1}^{A}\alpha_a(X,Y)\,\text{Anchor}_a(X,Y),
\]
with \(\text{Anchor}_a(X,Y)=Z_a\) the physical height of anchor \(a\).
We project \(\mathcal{H}_{\text{conf}}\) back to the image plane and compare it to a binary 2D ground mask.
This auxiliary loss penalises high confidence on non-ground regions—e.g.\ when a steep uphill anchor samples sky pixels—driving the network to deactivate such anchors.
As a result, \(\alpha_a\) becomes ground-consistent, and the ensuing height estimate is more reliable.

The resulting height-aware feature \(F_{\text{SA}}\) is finally fed to a lightweight decoder—%
a stack of multi-scale \(3 \times 3\) convolutional layers with residual connections—%
which outputs the BEV heightmap prediction.

\subsection{Height Consistency Module}
\label{sec:heightconsis}
\begin{figure}[t]
    \centering
    \includegraphics[width=1\linewidth]{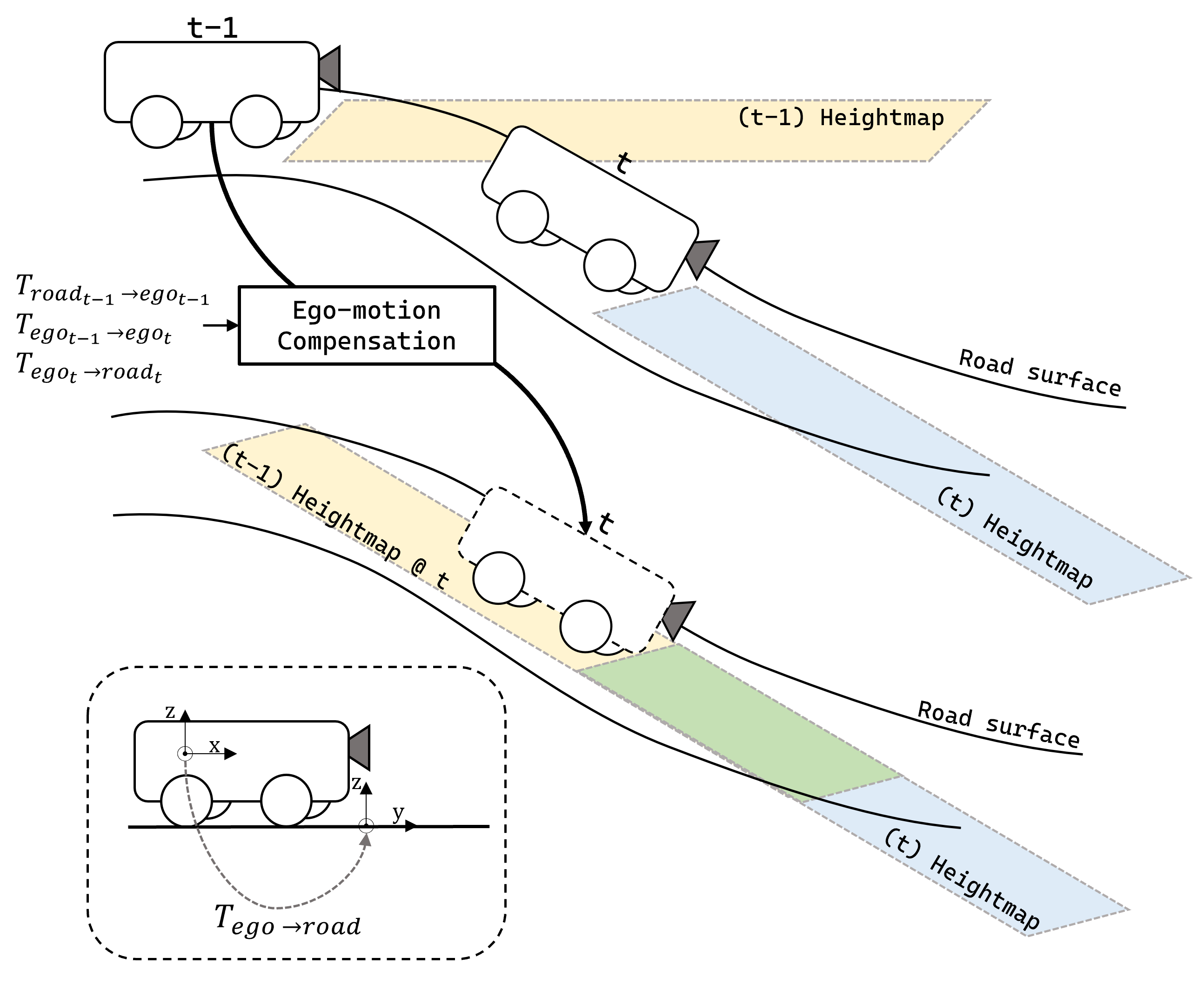}

   \caption{The illustration of the ego motion compensation from frame $t-1$ to frame $t$. The area shown in light green represents the region where the L1 loss is applied after the compensation. The dashed box represents the transformation from ego coordinate to road coordinate.}
   \label{fig:egomotion}
\end{figure}

Road surfaces are inherently static and continuous, meaning that height values in overlapping regions should remain consistent across different time steps, regardless of when they are predicted. To enforce this property, we introduce the Height Consistency Module, which ensures temporal stability in height estimation across consecutive frames.

As illustrated in \cref{fig:architecture}, the Height Consistency Module operates during the training stage by leveraging ego-motion compensation. The heightmap transformed from the front-view features at time step $t-1$ is aligned with time step $t$ through ego-motion compensation. This alignment allows the heightmap predicted at $t-1$ to be compared with the heightmap predicted from the current image features, enabling the computation of a height consistency loss. By enforcing consistency across consecutive frames, the model learns to mitigate temporal inconsistencies and generate more stable height predictions.

Further details of this process are depicted in \cref{fig:egomotion}. 
Given that our predicted heightmap \( \mathcal{H}\) is defined in the road coordinate system, we first establish the road plane $P_{\text{road}}$, which is tangent to the local road surface. 

Following \cite{Garnett20193dlanenet}, we define the road coordinates $C_{\text{road}} = (x,y,z)$ as follows:
The z-axis is aligned with the normal to $P_{\text{road}}$, while the y-axis corresponds to the projection of the camera's viewing direction, with the origin defined as the projection of the camera center onto $P_{\text{road}}$. Ego coordinates are aligned with respect to the vehicle’s rear axle, and the transform between ego and camera coordinate systems is provided by the known extrinsic calibration, allowing straightforward conversion.

To enforce temporal consistency between the heightmaps at time steps $t$ and $t-1$, we must align the two coordinate frames. This is achieved through the following transformations:
\[
\mathcal{H}_{\text{road}}^t (t-1) = T_{t-1}^{t} \mathcal{H}_{\text{road}}^{t-1} (t-1),
\]
\[
   T_{t-1}^{t} = T_{\text{ego} \to \text{road}}^t \cdot T_{\text{ego}(t-1) \to \text{ego}(t)} \cdot T_{\text{road} \to \text{ego}}^{t-1}.
\]

Thus, \( \mathcal{H}(t-1)\), the heightmap initially defined in road coordinates at $t-1$, is projected to the road coordinate system at $t$ as green area in \cref{fig:egomotion}. By applying an L1 loss to the regions that overlap between transformed
\( \mathcal{H}_{\text{road}}^t (t-1)\) and \( \mathcal{H}(t)\), we enforce temporal consistency across frames.

\subsection{Losses}
In the height estimation framework, we employ two types of loss functions.

To prevent the heightmap anchors from focusing on non-ground regions, we construct $\mathcal{H}_{conf}$ as a weighted sum of the heightmap anchors and their corresponding confidence values. We then project $\mathcal{H}_{conf}$ into the 2D image space using the camera extrinsic parameters $\mathbf{E}$ and intrinsic parameters $\mathbf{I}$. Finally, we compute the IoU loss between the projected $\mathcal{H}_{conf}$ and the 2D ground segmentation mask $\mathbf{M}_{ground}$  to guide slope-aware adaptive feature.
\[
\mathcal{L}_{SA} =  \text{IoU}(\mathbf{M}_{ground}, \mathbf{I} \cdot\mathbf{E} \cdot T_{\text{road} \to \text{ego}} \cdot \mathcal{H}_{conf}) 
\]

For the Consistent Loss, as mentioned in \cref{sec:heightconsis}, ego-motion compensation is performed, followed by applying L1 loss to the overlapping regions $\mathbf{M}_{overlapped}$ between time steps $t-1$ and $t$ to enforce temporal consistency.

\[
\mathcal{L}_{Cons} =  L1(\mathbf{M}_{overlapped}\mathcal{H}_{\text{road}}^t (t-1), \mathbf{M}_{overlapped}\mathcal{H}(t)) 
\]

The total loss used in the Height Estimation Framework is defined as follows:
\[
\mathcal{L}_{Height} =  \lambda_{SA}\mathcal{L}_{SA} + \lambda_{Cons}\mathcal{L}_{Cons} + \lambda_{h}L1(\mathcal{H}(t), \mathcal{H}_{gt}(t)) ,
\]
where $\mathcal{H}(t)$ is predicted heightmap at $t$ and $\mathcal{H}_{gt}(t)$ is ground truth heightmap at $t$. 
\section{Experiments}
\begin{figure*}[ht]
    \centering
    \includegraphics[width=1.0\linewidth]{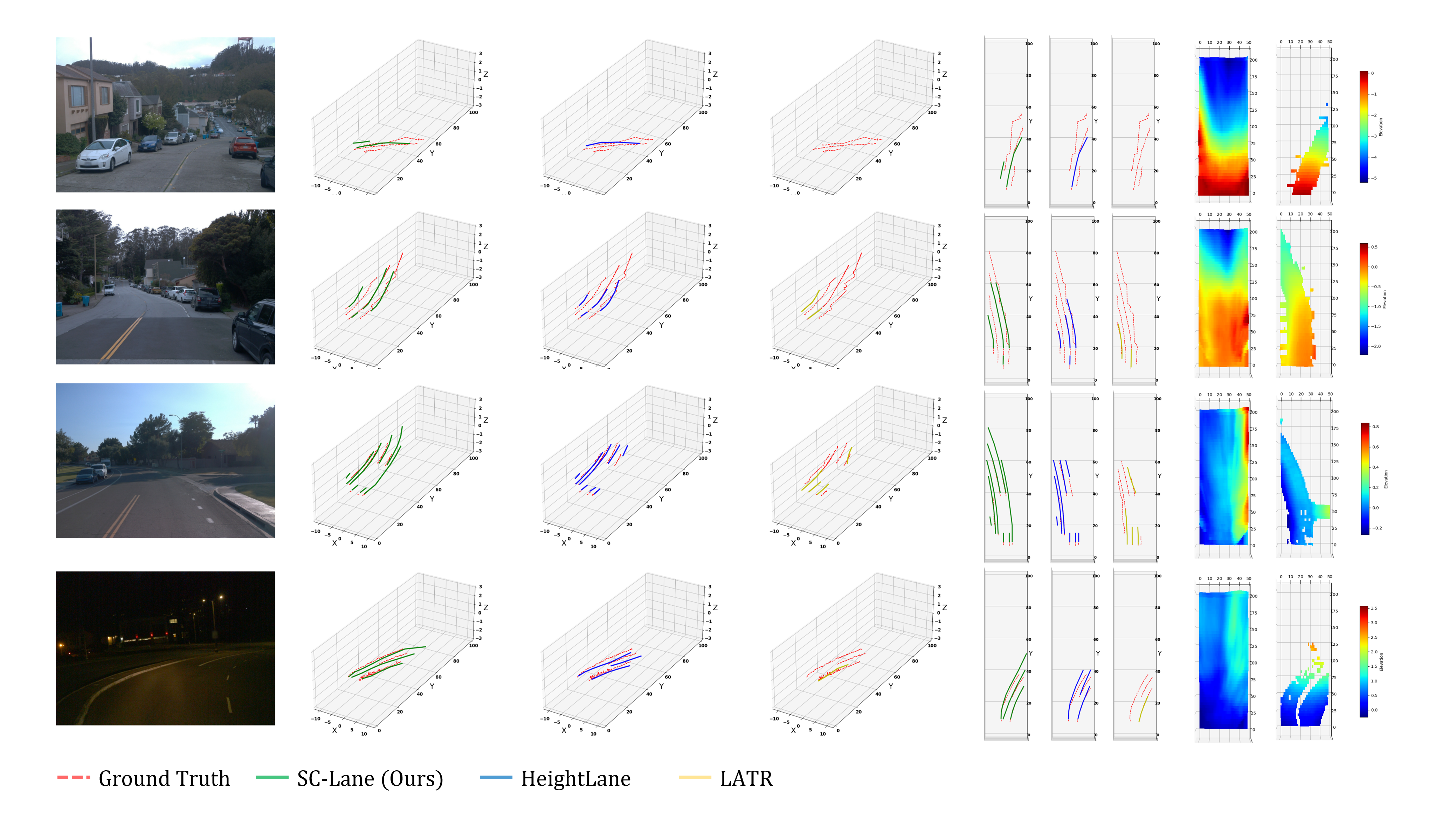}

\caption{\textbf{Qualitative Comparison of 3D Lane Detection.} Visualization of lane estimation results for SC-Lane, Ground Truth, and HeightLane\cite{park2024heightlane}. The first column presents input images from various road conditions. The next three columns display the corresponding 3D lane estimation results from different methods, with the ground truth shown in red, SC-Lane predictions in green, HeightLane predictions in blue, and LATR predictions in yellow. The following three columns show the BEV views of these results for clearer comparison. The last three columns present the predicted heightmap, the ground-truth heightmap, and the corresponding height colorbar, respectively.}

   \label{fig:qualitative}
\end{figure*}

\subsection{Implementation Details}
\paragraph{Dataset}
As proposed in \cite{park2024heightlane}, we generate heightmaps from the OpenLane \cite{chen2022persformer} dataset by accumulating ground-surface 3D point clouds using LiDAR data and segmentation labels. We define the heightmap with a resolution of 200×48 and a scale of 0.5 meters per pixel, matching the prediction range for 3D lane detection.

For the Apollo Synthetic dataset \cite{guo2020gen}, while LiDAR data is unavailable, certain subset scenarios provide depth maps and segmentation labels. Given the correlation between depth and height, we can easily derive the heightmap from the available depth information. Similarly, we define the Apollo dataset heightmap with a 200×48 resolution and 0.5 meters per pixel scale. Since not all frames in Apollo include semantic labels, we train and test only on the subset of frames with available segmentation annotations; the detailed results are provided in the supplementary material.

\paragraph{Training details}

We use ResNet-50 \cite{he2015deepresiduallearningimage} as the image feature extractor, with an input image size of 600 × 800. 
Training was conducted on four A6000 GPUs. For joint training of height estimation and lane detection, we trained for 24 epochs with a batch size of 8. When training height estimation alone, we used a batch size of 16 for 16 epochs.
The loss weights for the height estimation framework are set to $\lambda_{SA}$ = 5, $\lambda_{Cons}$ = 2, $\lambda_{h}$ = 10.

\subsection{Metrics}

In this section, we propose height estimation metrics to quantitatively evaluate the accuracy of road height predictions. Since road height estimation differs from depth estimation in terms of scale and characteristics, we employ absolute error-based metrics and threshold-based accuracy inspired by depth estimation evaluation.

The mean absolute error (MAE) is defined as
\[
\text{MAE} = \frac{1}{N} \sum_{i=1}^{N} \left|\, \mathcal{H}_i^{\text{pred}} - \mathcal{H}_i^{\text{gt}} \,\right|,
\]
which measures the average absolute difference between the predicted height $h_i^{\text{pred}}$ and the ground truth height $h_i^{\text{gt}}$. A lower MAE indicates more precise height predictions.

The root mean square error (RMSE) is given by
\[
\text{RMSE} = \sqrt{\frac{1}{N} \sum_{i=1}^{N} \left( \mathcal{H}_i^{\text{pred}} - \mathcal{H}_i^{\text{gt}} \right)^2 },
\]
which penalizes large errors more than MAE by squaring the differences, making it more sensitive to outliers.

To assess the percentage of predictions that fall within an acceptable error range, we introduce a threshold-based accuracy metric:
\[
Acc@t = \frac{1}{N} \sum_{i=1}^{N} \mathbf{1} \left( \left|\, \mathcal{H}_i^{\text{pred}} - \mathcal{H}_i^{\text{gt}} \,\right| < t \right),
\]
where $t \in \{0.05m, 0.1m, 0.2m\}$ and $\mathbf{1}(\cdot)$ is an indicator function that returns 1 if the condition is met and 0 otherwise. Inspired by the commonly used delta accuracy ($\delta$) in depth estimation, this metric provides an intuitive measure of how well the predicted heights match the ground truth within reasonable error margins. However, unlike depth values, which can vary significantly across a scene, road height values are relatively constrained. As a result, absolute error thresholds serve as a more practical choice for evaluating height estimation performance in this context.

\subsection{Qualitative Result}
In \cref{fig:qualitative}, we compare the lane estimation performance of SC-Lane against Ground Truth, HeightLane\cite{park2024heightlane} and LATR\cite{luo2023latr} in both 3D and BEV views. The figure includes various challenging scenarios such as sloped road, curved path, and low-light condition, illustrating the robustness of our method. Unlike HeightLane and LATR, which struggles with lane misalignment and height inconsistencies, SC-Lane maintains accurate lane positioning and structural consistency. Moreover, we visualize the predicted heightmap alongside the ground-truth heightmap, showing that SC-Lane's height estimation framework successfully captures the lane elevation with high fidelity. The 3D visualizations of these heightmap results are provided in the supplementary material for further reference. These results validate the effectiveness of our approach in handling complex real-world driving scenarios.

\begin{table*}
  \centering 
  \begin{tabular}{@{}l|ccccc@{}}
    \hline
    \textbf{Method} & \textbf{F-score(\%)} & \textbf{X-error (near)} & \textbf{X-error (far)} & \textbf{Z-error (near)} & \textbf{Z-error (far)} \\
    \hline
    PersFormer\cite{chen2022persformer} & 50.5 &0.485 &0.553& 0.364 &0.431\\
    Anchor3DLane\cite{huang2023anchor3dlane} & 53.1 & 0.300 &0.311 &0.103 &0.139\\
    BEV-LaneDet\cite{wang2023bev} & 58.4 &0.309& 0.659 &0.244 &0.631 \\
    LaneCPP\cite{pittner2024lanecpp} &60.3& 0.264 & 0.310 &0.077 & \underline{0.117}\\
    LATR\cite{luo2023latr} & 61.9 &\textbf{0.219} &\underline{0.259} &\underline{0.075}& \textbf{0.104} \\
    HeightLane\cite{park2024heightlane} & \underline{62.7} & 0.240 & 0.266 &0.116& 0.165\\
    RFTR \cite{li2024repvf} & 61.8 & 0.341 & 0.450 & \textbf{0.073}&  0.107\\
    PVALane\cite{zheng2024pvalane} & \underline{62.7} & 0.232 & 0.259 & 0.092& 0.118\\
    \OURS(ours) & \textbf{64.3} & \underline{0.227} & \textbf{0.251} &0.088&0.128\\
    \hline
  \end{tabular}
  \caption{Quantitative results comparison with other models on the OpenLane validation set. The best results are highlighted in \textbf{bold} and second-best results are \underline{underlined}.}
  \label{tab:metric2}
  \vspace{-0.1em}
\end{table*}

\begin{table*}
    \centering 
    \begin{tabular}{l|ccccccc}
    \hline
    \textbf{Method} & \textbf{All} & \textbf{Up \& Down} & \textbf{Curve} & \textbf{Extreme Weather} & \textbf{Night} & \textbf{Intersection} & \textbf{Merge \& Split} \\ \hline
    BEVLaneDet\cite{wang2023bev}      & 58.4         & 48.7                & 63.1           & 53.4                     & 53.4           & 50.3                  & 53.7                    \\
    LATR\cite{luo2023latr}            & 61.9         & \textbf{55.2}                & 68.2           & 57.1                     & 55.4           & 52.3                  & \underline{61.5}                    \\
    HeightLane\cite{park2024heightlane}      & \underline{62.7}         & 53.6                & \underline{69.3}           & 55.4                     & 54.6           & \underline{54.1}                  & 61.1                    \\
    PVALane\cite{zheng2024pvalane}         & \underline{62.7}         & 54.1                & 67.3           & \textbf{62.0}                     & \textbf{57.2}           & 53.4                  & 60.0                    \\
    \OURS(ours)    & \textbf{64.3}        & \underline{54.6}                & \textbf{70.7}           & \underline{58.1}                     & \underline{56.5}           & \textbf{56.1}                  & \textbf{63.0}                    \\ \hline
    \end{tabular}
    \caption{Quantitative results comparison by scenario on the OpenLane validation set using F-score. The best results for each scenario are highlighted in \textbf{bold} and second-best results are \underline{underlined}.}
    \label{tab:scenario2}
\end{table*}

\vspace{-0.3em}
\subsection{Quantitative Result}

We evaluate SC-Lane on both the OpenLane\cite{chen2022persformer} and ApolloSim \cite{guo2020gen}. The results on Apollo are provided in the supplementary materials, while this section focuses on OpenLane.

\paragraph{Overall Performance Comparison} \cref{tab:metric2} presents a quantitative comparison of our method, \OURS, against existing state-of-the-art approaches on the OpenLane validation set. Our method achieved the highest F-score of 64.3\%, outperforming previous approaches. The improvement in F-score demonstrates the effectiveness of our suggested modules in refining lane predictions by ensuring a more stable and accurate height estimation.
On the other hand, SC-Lane did not achieve strong performance in Z-error, which is closely related to the representation used in different lane estimation methods.
BEVLaneDet \cite{wang2023bev}, HeightLane\cite{park2024heightlane}, and SC-Lane utilize dense BEV features to generate segmentation maps, offset maps, and embedding maps in the BEV space, with loss computation also performed on these BEV representations. In contrast, methods such as LATR\cite{luo2023latr} and LaneCPP\cite{pittner2024lanecpp} predefine longitudinal lane anchors (e.g., 5m, 10m, 20m, ... 100m) or control points and predict the X and Z coordinates at these anchor points, resulting in loss computation directly in 3D space.
This difference significantly impacts Z-error metrics, and while neither approach is inherently superior, our method prioritizes dense heightmaps, which naturally aligns with dense BEV features. As a result, we accept a slight trade-off in Z-error performance in favor of a representation more suitable for our framework.

\paragraph{Performance Analysis by Scenario} Beyond the overall results on OpenLane, we further evaluate our model across various challenging scenarios within the OpenLane validation set. As shown in \cref{tab:scenario2}, an in-depth evaluation of SC-Lane demonstrates its consistently strong performance across all driving environments. Notably, \OURS achieves the highest performance in curved roads, merge \& split scenarios, and intersections, outperforming existing state-of-the-art methods. It records an F-score of 70.7\% in curved roads, 63.0\% in merge \& split scenarios, and 56.1\% in intersections, highlighting its effectiveness in handling complex lane structures.

Additionally, \OURS achieves the second-best performance in up \& down terrain, extreme weather, and night conditions. It attains an F-score of 54.6\% in up \& down terrain, 58.1\% in extreme weather, and 56.5\% in night scenarios, demonstrating its ability to maintain stable performance under varying lighting and environmental conditions.

Unlike previous models that excel only in specific scenarios, \OURS consistently delivers robust performance across all driving environments. This indicates that the model is not overfitting to particular conditions but rather maintains strong generalization capabilities across diverse road and weather settings. These results establish \OURS as a highly reliable solution for lane detection, capable of handling complex road structures, varying lighting conditions, and adverse weather, making it a dependable choice for real-world deployment.

\vspace{-0.5cm}
\paragraph{Quantitative Comparison using Heightmap Estimation}
\begin{table}[h]
    \small
    \begin{tabular}{c|cc|ccc}
        \hline
        \multirow{2}{*}{\textbf{Method}} 
            & \multicolumn{2}{c|}{\textbf{Error Metrics} ($\downarrow$)} 
            & \multicolumn{3}{c}{\textbf{Accuracy Metrics} ($\uparrow$)} \\
        \cline{2-6}
            & MAE & RMSE & @0.05 & @0.1 & @0.2 \\
        \hline
        Single Anchor   & 0.234 & 0.367 & 0.253 & 0.452 & 0.688 \\
        HeightLane      & 0.235 & 0.343 & 0.253 & 0.442 & 0.673 \\
        \OURS           & \textbf{0.176} & \textbf{0.259} & \textbf{0.293} & \textbf{0.507} & \textbf{0.756} \\
        \hline
    \end{tabular}
    \caption{Comparison of height estimation performance across models using our proposed heightmap estimation metric.}
    \label{tab:height_estimation_heightmap_only}
\end{table}

\begin{table*}[h]
    \centering
    \begin{tabular}{l|cc|ccc|c}
        \hline
        \multirow{2}{*}{\textbf{Method}} 
            & \multicolumn{2}{c|}{\textbf{Error Metrics} ($\downarrow$)} 
            & \multicolumn{3}{c|}{\textbf{Accuracy Metrics} ($\uparrow$)}
            & \multirow{2}{*}{\textbf{F-score} ($\uparrow$)} \\
        \cline{2-6}
            & MAE & RMSE & Acc@0.05 & Acc@0.1 & Acc@0.2 & \\
        \hline
        Baseline (HeightLane)               & 0.236 & 0.368 & 0.247 & 0.423 & 0.647 & 62.7 \\
        + Height Consistency                & 0.228 & 0.349 & 0.252 & 0.459 & 0.688 & 63.2 \\
        + SAA Feature\textsuperscript{*}    & 0.219 & 0.335 & 0.261 & 0.463 & 0.712 & 63.9 \\
        + SAA Feature                       & 0.216 & 0.321 & 0.268 & 0.472 & 0.718 & 64.1 \\
        All                                 & \textbf{0.198} & \textbf{0.301} & \textbf{0.282} & \textbf{0.493} & \textbf{0.739} & \textbf{64.3} \\
        \hline
    \end{tabular}
    \caption{Comparison of height estimation performance with different enhancements applied to the baseline, where SAA denotes the Slope-Aware Adaptive Feature and the superscript \textsuperscript{*} indicates the variant trained without auxiliary ground mask supervision.}
    \label{tab:height_estimation_variants}
\end{table*}

\cref{tab:height_estimation_heightmap_only} presents the performance of various models evaluated using our newly proposed heightmap estimation metric. This metric provides a unified and quantitative way to assess the accuracy of height estimation modules, enabling, for the first time, direct comparison across previously disparate approaches. Our model \OURS significantly outperforms existing baselines—Single Anchor and HeightLane—achieving a MAE of 0.176 and RMSE of 0.259, along with consistently higher accuracy across all thresholds (Acc@0.05 / 0.1 / 0.2). These improvements suggest that generating more informative height features via our Multi-Slope Anchors and Slope-Aware Adaptive feature, together with enforcing temporal coherence through our height consistency loss across consecutive frames, jointly contribute to the more precise and robust height estimation achieved by \OURS.

\subsection{Ablation Study}

\paragraph{Impact of Each Height Estimation Framework Module}
Through \cref{tab:height_estimation_variants}, we analyze the impact of Slope-Aware Adaptive Feature and Height Consistency Module on the model's performance.

The Slope-Aware Adaptive Feature (SAA) assigns a learned weight $\alpha_a$ to each anchor, guiding the model to focus on the most relevant slopes. When trained without auxiliary ground mask supervision, SAA still learns these weights automatically and yields a 1.2-point improvement over the baseline. However, since the model lacks explicit guidance on which regions to attend to, we introduce an auxiliary loss using a ground segmentation mask; with this supervision, the SAA feature delivers a 1.4 point gain over the baseline. Overall, SAA substantially improves all metrics, with the largest boost seen in Acc@0.2 demonstrating its effectiveness in handling pronounced slope variations.

On the other hand, the Height Consistency Module alone did not yield substantial performance improvements. This is likely because vehicle motion and slope changes between consecutive frames are typically minimal, making the consistency constraint less impactful on its own. However, when both the Slope-Aware Adaptive Feature and the Height Consistency Module were used together, the model achieved significant improvements across all metrics, demonstrating their complementary effect in enhancing height estimation.

\begin{table}[h]
\centering
\small
\begin{tabular}{lccc}
\hline
\textbf{Method} & \textbf{Frames} & \textbf{F-Score} ↑  & \textbf{Temporal cue} \\
\hline
Anchor3DLane \cite{huang2023anchor3dlane}     & 2 & 54.3 & Input \\ 
Curveformer++ \cite{bai2025curveformer3dlanedetection}  & 2 & 55.4 & Input \\
Curveformer++ \cite{bai2025curveformer3dlanedetection} & 3 & 53.5 & Input \\ 
GTANet \cite{zheng2025geometry}                         & 2 & 62.4 & Input \\ 
\textbf{Ours (SC-Lane)}                       & 2 & \textbf{64.3} & Supervision \\
\bottomrule
\end{tabular}
\vspace{-3.1mm}
\caption{Comparison with multi-frame 3D lane detection methods. “Temporal cue” indicates whether temporal cues are fed as input at inference (“Input”) or only used as training supervision (“Supervision”) with no extra inference cost.}
\label{tab:multi-frame}
\end{table}

\paragraph{Comparison with Multi-Frame 3D Lane Detection Methods}
We compare SC-Lane against state-of-the-art multi-frame models in \cref{tab:multi-frame}. Although our training leverages two consecutive frames, inference still requires only a single frame—this recipe can be applied to any single-frame model at no extra inference cost. Even with single-frame inference, SC-Lane outperforms all compared multi-frame approaches by a substantial margin. Applied to HeightLane, our two-frame training raises its F-score from 62.7 to 63.2 (+0.5) as mentioned in \cref{tab:height_estimation_variants}.

\section{Conclusion} 
In this paper, we introduced SC-Lane, a slope-aware and temporally consistent height estimation framework for 3D lane detection. By leveraging the Slope-Aware Adaptive Feature module for dynamic slope fusion and the Height Consistency Module for enforcing temporal stability, SC-Lane significantly improves height estimation accuracy and robustness.
Experiments demonstrate that SC-Lane achieves state-of-the-art performance in 3D lane detection (64.3\% F-score on OpenLane benchmark) and establishes strong baseline results for height estimation, where it introduces new evaluation metrics and outperforms existing models in these metrics. While designed for lane detection, its height estimation framework can be extended to HD map reconstruction, localization, and other autonomous driving tasks.  

\section{Acknowledgement}
This work was supported by grants from KEIT (No. 1415181767, Development of a multi-angle polarized camera), NRF (No. RS-2024-00359718), and IITP (No. RS-2021-II211343, AI Graduate School Program at Seoul National University), funded by the Korea government (MOTIE/MSIT).

{
    \small
    \bibliographystyle{ieeenat_fullname}

\begin{thebibliography}{30}
        \providecommand{\natexlab}[1]{#1}
        \providecommand{\url}[1]{\texttt{#1}}
        \expandafter\ifx\csname urlstyle\endcsname\relax
          \providecommand{\doi}[1]{doi: #1}\else
          \providecommand{\doi}{doi: \begingroup \urlstyle{rm}\Url}\fi
        
        \bibitem[Bai et~al.(2025)]{bai2025curveformer3dlanedetection}
        Yifeng Bai et~al.
        \newblock curveformer++:3d lane detection by curve propagation with temporal curve query and attention, 2025.
        
        \bibitem[Chen et~al.(2022)Chen, Sima, Li, Zheng, Xu, Geng, Li, He, Shi, Qiao, et~al.]{chen2022persformer}
        Li Chen, Chonghao Sima, Yang Li, Zehan Zheng, Jiajie Xu, Xiangwei Geng, Hongyang Li, Conghui He, Jianping Shi, Yu Qiao, et~al.
        \newblock Persformer: 3d lane detection via perspective transformer and the openlane benchmark.
        \newblock In \emph{European Conference on Computer Vision}, pages 550--567. Springer, 2022.
        
        \bibitem[Chung et~al.(2024)Chung, Georgakis, Spieler, Padgett, Agha, and Khattak]{chung2024pixel}
        Chanyoung Chung, Georgios Georgakis, Patrick Spieler, Curtis Padgett, Ali Agha, and Shehryar Khattak.
        \newblock Pixel to elevation: Learning to predict elevation maps at long range using images for autonomous offroad navigation.
        \newblock \emph{IEEE Robotics and Automation Letters}, 2024.
        
        \bibitem[Ding et~al.(2024)Ding, Jiang, Li, Chen, and Huang]{ding2024towards}
        Laiyan Ding, Hualie Jiang, Jie Li, Yongquan Chen, and Rui Huang.
        \newblock Towards cross-view-consistent self-supervised surround depth estimation.
        \newblock In \emph{2024 IEEE/RSJ International Conference on Intelligent Robots and Systems (IROS)}, pages 10043--10050. IEEE, 2024.
        
        \bibitem[Garnett et~al.(2019)Garnett, Cohen, Pe'er, Lahav, and Levi]{Garnett20193dlanenet}
        Noa Garnett, Rafi Cohen, Tomer Pe'er, Roee Lahav, and Dan Levi.
        \newblock 3d-lanenet: End-to-end 3d multiple lane detection.
        \newblock In \emph{Proceedings of the IEEE/CVF International Conference on Computer Vision (ICCV)}, 2019.
        
        \bibitem[Godard et~al.(2017)Godard, Aodha, and Brostow]{8100182}
        Clément Godard, Oisin~Mac Aodha, and Gabriel~J. Brostow.
        \newblock Unsupervised monocular depth estimation with left-right consistency.
        \newblock In \emph{2017 IEEE Conference on Computer Vision and Pattern Recognition (CVPR)}, pages 6602--6611, 2017.
        
        \bibitem[Godard et~al.(2019)Godard, Mac~Aodha, Firman, and Brostow]{godard2019digging}
        Cl{\'e}ment Godard, Oisin Mac~Aodha, Michael Firman, and Gabriel~J. Brostow.
        \newblock Digging into self-supervised monocular depth estimation.
        \newblock In \emph{Proceedings of the IEEE/CVF International Conference on Computer Vision}, pages 3828--3838, 2019.
        
        \bibitem[Guo et~al.(2020)Guo, Chen, Zhao, Zhang, Miao, Wang, and Choe]{guo2020gen}
        Yuliang Guo, Guang Chen, Peitao Zhao, Weide Zhang, Jinghao Miao, Jingao Wang, and Tae~Eun Choe.
        \newblock Gen-lanenet: A generalized and scalable approach for 3d lane detection.
        \newblock In \emph{Computer Vision--ECCV 2020: 16th European Conference, Glasgow, UK, August 23--28, 2020, Proceedings, Part XXI 16}, pages 666--681. Springer, 2020.
        
        \bibitem[He et~al.(2015)He, Zhang, Ren, and Sun]{he2015deepresiduallearningimage}
        Kaiming He, Xiangyu Zhang, Shaoqing Ren, and Jian Sun.
        \newblock Deep residual learning for image recognition, 2015.
        
        \bibitem[Hou et~al.(2023)Hou, Gan, and Yokoya]{hou2023monocular}
        X. Hou, W. Gan, and N. Yokoya.
        \newblock Enhancing monocular height estimation from aerial images with street-view images.
        \newblock \emph{arXiv preprint}, arXiv:2311.02121, 2023.
        
        \bibitem[Hu et~al.(2024)Hu, Gao, Li, Zhao, Cun, Zhang, and Shan]{hu2024depthcrafter}
        Wenbo Hu, Xiaoyi Gao, Xinyu Li, Shengnan Zhao, Xiaodong Cun, Yizhi Zhang, and Ying Shan.
        \newblock Depthcrafter: Generating consistent long depth sequences for open-world videos.
        \newblock \emph{arXiv preprint arXiv:2409.02095}, 2024.
        
        \bibitem[Huang et~al.(2023)]{huang2023anchor3dlane}
        Shaofei Huang et~al.
        \newblock Anchor3dlane: Learning to regress 3d anchors for monocular 3d lane detection.
        \newblock In \emph{CVPR}, pages 17451--17460, 2023.
        
        \bibitem[Ke et~al.(2024)Ke, Narnhofer, Huang, Ke, Peters, Fragkiadaki, Obukhov, and Schindler]{ke2024rollingdepth}
        Bingxin Ke, Dominik Narnhofer, Shengyu Huang, Lei Ke, Torben Peters, Katerina Fragkiadaki, Anton Obukhov, and Konrad Schindler.
        \newblock Video depth without video models, 2024.
        
        \bibitem[Li et~al.(2024)Li, Han, Yin, Zhao, and Shen]{li2024repvf}
        Chunliang Li, Wencheng Han, Junbo Yin, Sanyuan Zhao, and Jianbing Shen.
        \newblock Repvf: A unified vector fields representation for multi-task 3d perception.
        \newblock In \emph{European Conference on Computer Vision}, pages 273--292. Springer, 2024.
        
        \bibitem[Li et~al.(2020)Li, Wang, and Fang]{li2020height}
        Xiang Li, Mingyang Wang, and Yi Fang.
        \newblock Height estimation from single aerial images using a deep ordinal regression network.
        \newblock \emph{IEEE Geoscience and Remote Sensing Letters}, 19:\penalty0 1--5, 2020.
        
        \bibitem[Liu et~al.(2023)Liu, Yan, Jia, Li, Gao, Wang, and Zhang]{liu2023petrv2}
        Yingfei Liu, Junjie Yan, Fan Jia, Shuailin Li, Aqi Gao, Tiancai Wang, and Xiangyu Zhang.
        \newblock Petrv2: A unified framework for 3d perception from multi-camera images.
        \newblock In \emph{Proceedings of the IEEE/CVF International Conference on Computer Vision}, pages 3262--3272, 2023.
        
        \bibitem[Luo et~al.(2023)Luo, Zheng, Yan, Kun, Zheng, Cui, and Li]{luo2023latr}
        Yueru Luo, Chaoda Zheng, Xu Yan, Tang Kun, Chao Zheng, Shuguang Cui, and Zhen Li.
        \newblock Latr: 3d lane detection from monocular images with transformer.
        \newblock In \emph{Proceedings of the IEEE/CVF International Conference on Computer Vision}, pages 7941--7952, 2023.
        
        \bibitem[Moon et~al.(2024)Moon, Bello, Kwon, and Kim]{moon2024from}
        Jisoo Moon, Jose L.~Gonzalez Bello, Byeongjun Kwon, and Minsu Kim.
        \newblock From-ground-to-objects: Coarse-to-fine self-supervised monocular depth estimation of dynamic objects with ground contact prior.
        \newblock In \emph{Proceedings of the IEEE/CVF Conference on Computer Vision and Pattern Recognition}, pages 10519--10529, 2024.
        
        \bibitem[Mou and Zhu(2018)]{mou2018im2height}
        Lichao Mou and Xiao~Xiang Zhu.
        \newblock Im2height: Height estimation from single monocular imagery via fully residual convolutional-deconvolutional network.
        \newblock \emph{arXiv preprint arXiv:1802.10249}, 2018.
        
        \bibitem[Park et~al.(2025)Park, Seo, and Lim]{park2024heightlane}
        Chaesong Park, Eunbin Seo, and Jongwoo Lim.
        \newblock Heightlane: Bev heightmap guided 3d lane detection.
        \newblock In \emph{Proceedings of the Winter Conference on Applications of Computer Vision (WACV)}, pages 1692--1701, 2025.
        
        \bibitem[Pittner et~al.(2024)Pittner, Janai, and Condurache]{pittner2024lanecpp}
        Maximilian Pittner, Joel Janai, and Alexandru~P Condurache.
        \newblock Lanecpp: Continuous 3d lane detection using physical priors.
        \newblock In \emph{Proceedings of the IEEE/CVF Conference on Computer Vision and Pattern Recognition}, pages 10639--10648, 2024.
        
        \bibitem[Qiu et~al.(2024)Qiu, Pang, Fang, and Xue]{qiu2024heightmapnet}
        W. Qiu, S. Pang, J. Fang, and J. Xue.
        \newblock Heightmapnet: Explicit height modeling for end-to-end hd map learning.
        \newblock \emph{arXiv preprint}, arXiv:2411.01408, 2024.
        
        \bibitem[Wang et~al.(2023)Wang, Qin, Li, Li, Cao, and Xu]{wang2023bev}
        Ruihao Wang, Jian Qin, Kaiying Li, Yaochen Li, Dong Cao, and Jintao Xu.
        \newblock Bev-lanedet: An efficient 3d lane detection based on virtual camera via key-points.
        \newblock In \emph{Proceedings of the IEEE/CVF Conference on Computer Vision and Pattern Recognition}, pages 1002--1011, 2023.
        
        \bibitem[Wu et~al.(2024)Wu, Li, Qin, Zhao, and Li]{wu2024heightformer}
        Y. Wu, R. Li, Z. Qin, X. Zhao, and X. Li.
        \newblock Heightformer: Explicit height modeling without extra data for camera-only 3d object detection in bird’s eye view.
        \newblock \emph{IEEE Transactions on Image Processing}, 2024.
        
        \bibitem[Yang et~al.(2022)Yang, Zhang, Geng, Wang, and Liu]{yang2022real}
        Bowen Yang, Qingwen Zhang, Ruoyu Geng, Lujia Wang, and Ming Liu.
        \newblock Real-time neural dense elevation mapping for urban terrain with uncertainty estimations.
        \newblock \emph{IEEE Robotics and Automation Letters}, 8\penalty0 (2):\penalty0 696--703, 2022.
        
        \bibitem[Zhang et~al.(2019)Zhang, Shen, Li, Cao, Liu, and Yan]{zhang2019exploiting}
        Hao Zhang, Chunhua Shen, Yifan Li, Yiming Cao, Yifan Liu, and Yan Yan.
        \newblock Exploiting temporal consistency for real-time video depth estimation.
        \newblock In \emph{Proceedings of the IEEE/CVF International Conference on Computer Vision}, pages 1725--1734, 2019.
        
        \bibitem[Zhang et~al.(2025{\natexlab{a}})Zhang, Wei, Xu, Zhang, Wang, Zhang, and Liu]{zhang2025ts}
        Song Zhang, Zhiwei Wei, Wenjia Xu, Lili Zhang, Yang Wang, Jinming Zhang, and Junyi Liu.
        \newblock Ts-satmvsnet: Slope aware height estimation for large-scale earth terrain multi-view stereo.
        \newblock \emph{arXiv preprint arXiv:2501.01049}, 2025{\natexlab{a}}.
        
        \bibitem[Zhang et~al.(2025{\natexlab{b}})Zhang, Wei, Xu, Zhang, Wang, Zhang, and Liu]{zhang2025tssatmvsnet}
        S. Zhang, Z. Wei, W. Xu, L. Zhang, Y. Wang, J. Zhang, and J. Liu.
        \newblock Ts-satmvsnet: Slope aware height estimation for large-scale earth terrain multi-view stereo.
        \newblock \emph{arXiv preprint}, arXiv:2501.01049, 2025{\natexlab{b}}.
        
        \bibitem[Zheng et~al.(2025)Zheng, Han, Yan, Xu, and Shen]{zheng2025geometry}
        Huan Zheng, Wencheng Han, Tianyi Yan, Cheng-zhong Xu, and Jianbing Shen.
        \newblock Geometry-aware temporal aggregation network for monocular 3d lane detection.
        \newblock \emph{arXiv preprint arXiv:2504.20525}, 2025.
        
        \bibitem[Zheng et~al.(2024)Zheng, Zhang, Mou, Gao, Li, Huang, Pun, and Yuan]{zheng2024pvalane}
        Zewen Zheng, Xuemin Zhang, Yongqiang Mou, Xiang Gao, Chengxin Li, Guoheng Huang, Chi-Man Pun, and Xiaochen Yuan.
        \newblock Pvalane: prior-guided 3d lane detection with view-agnostic feature alignment.
        \newblock In \emph{Proceedings of the AAAI Conference on Artificial Intelligence}, pages 7597--7604, 2024.
        
    \end{thebibliography}

}
\end{document}